\documentclass[journal]{IEEEtran}
\usepackage{cite}
\usepackage{amsmath,amssymb,amsfonts}
\usepackage{graphicx}
\usepackage{textcomp}
\usepackage{multirow}

\usepackage{booktabs} 
\usepackage{tabularx}
\usepackage{graphicx} 
\usepackage{multirow}
\usepackage{algorithm,algpseudocode}
\usepackage{amsmath}

\def\BibTeX{{\rm B\kern-.05em{\sc i\kern-.025em b}\kern-.08em
		T\kern-.1667em\lower.7ex\hbox{E}\kern-.125emX}}
        
\begin{document}
	\title{Few-Shot Adaptation of Training-Free Foundation Model for 3D Medical Image Segmentation}
	\author{Xingxin He, Yifan Hu, Zhaoye Zhou, Mohamed Jarraya, and Fang Liu
		\thanks{Xingxin He, Yifan Hu, and Fang Liu are with the Athinoula A. Martinos Center for Biomedical Imaging, Massachusetts General Hospital, Charlestown, Massachusetts, USA and Harvard Medical School, Boston, Massachusetts, USA (email: xihe2@mgh.harvard.edu; yihu2@mgh.harvard.edu; fliu12@mgh.harvard.edu).}
		\thanks{Zhaoye Zhou and Mohamed Jarraya are with the Department of Radiology, Massachusetts General Hospital, Harvard Medical School, Boston, Massachusetts, USA (email: zzhou27@mgh.harvard.edu; mjarraya@mgh.harvard.edu)}}

	\maketitle
	
	\begin{abstract}
        Vision foundation models have achieved remarkable progress across various image analysis tasks. In the image segmentation task, foundation models like the Segment Anything Model (SAM) enable generalizable zero-shot segmentation through user-provided prompts. However, SAM primarily trained on natural images, lacks the domain-specific expertise of medical imaging. This limitation poses challenges when applying SAM to medical image segmentation, including the need for extensive fine-tuning on specialized medical datasets and a dependency on manual prompts, which are both labor-intensive and require intervention from medical experts.

        This work introduces the Few-shot Adaptation of Training-frEe SAM (FATE-SAM), a novel method designed to adapt the advanced Segment Anything Model 2 (SAM2) for 3D medical image segmentation. FATE-SAM reassembles pre-trained modules of SAM2 to enable few-shot adaptation, leveraging a small number of support examples to capture anatomical knowledge and perform prompt-free segmentation, without requiring model fine-tuning. To handle the volumetric nature of medical images, we incorporate a Volumetric Consistency mechanism that enhances spatial coherence across 3D slices. We evaluate FATE-SAM on multiple medical imaging datasets and compare it with supervised learning methods, zero-shot SAM approaches, and fine-tuned medical SAM methods. Results show that FATE-SAM delivers robust and accurate segmentation while eliminating the need for large annotated datasets and expert intervention. FATE-SAM provides a practical, efficient solution for medical image segmentation, making it more accessible for clinical applications.
		
	\end{abstract}
	
	\vspace{1.2cm}
	
	\section{Introduction}
	\label{sec:introduction}
	\IEEEPARstart{I}{mage} segmentation is an important step in medical image analysis, as it can assist in downstream applications such as disease diagnosis, treatment planning, and monitoring of disease progression \cite{ma2024segment}. Deep learning approaches have shown significant potential in advancing medical image segmentation \cite{wang2022medical}. However, supervised deep learning methods are inherently data-driven, relying heavily on large-scale labeled datasets for training. This dependence presents considerable challenges in the medical field, where manual annotation requires specialized expertise and is both labor-intensive and costly.
	
	The emergence of vision foundation models, pre-trained on vast and diverse datasets to capture generalized image features, offers a promising alternative to address downstream tasks with limited labeled data and minimal modifications to the pre-trained models \cite{radford2021learning}. Dedicated segmentation foundation models, such as the Segment Anything Model (SAM) \cite{kirillov2023segment}, trained on over 11 million static images and 1 billion masks, and its advanced successor SAM2 \cite{ravi2024sam}, trained on 51,000 videos and 600,000 time-series masks, enable zero-shot inference for general image segmentation using user-provided prompts, such as points, bounding boxes, or masks, to indicate the target object for segmentation.
	
	In medical imaging, pre-trained SAM has been utilized as zero-shot predictors in some recent studies\cite{mazurowski2023segment}. However, transitioning from the natural image domain to medical image segmentation presents notable challenges due to the significant differences between these two domains. Medical images often contain indistinct boundaries, complex textures, and specific anatomical details that differ greatly from the clear object features typically found in natural images. These complexities can make medical image segmentation difficult through zero-shot inference, even with exact manual prompts provided \cite{zhang2024segment}. To bridge this domain gap, several studies have focused on fine-tuning SAM using large, curated medical image datasets to incorporate domain-specific knowledge \cite{ma2024segment,wang2024sammed3dgeneralpurposesegmentationmodels}. While fine-tuning can enhance performance, it also introduces additional computational costs and highlights the need for extensive annotated datasets. Furthermore, both zero-shot inference and the fine-tuned medical SAM rely on user-provided prompts for segmentation which can be labor-intensive and requires medical expert intervention.
	
	\begin{figure*}[ht!]
		\centering
		\vspace{-0.7cm}
		\includegraphics[width=0.96\textwidth]{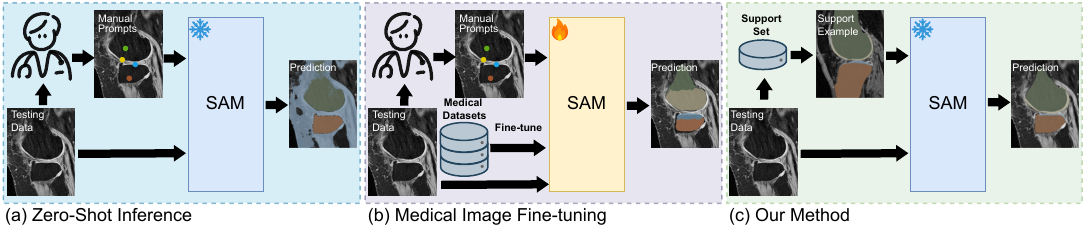}
		\vspace{-0.4cm}
		\caption{Applications of SAM for medical image segmentation:
            a) \textbf{Zero-Shot Inference}: Relies on manual prompts and struggles with complex anatomical structures due to the absence of medical domain-specific training.
			b) \textbf{Medical Image Fine-Tuning}: Enhances domain adaptation by fine-tuning on medical datasets. However, it demands extensive data collection and also relies on manual prompts.
			c) \textbf{Our Method}: A training-free and prompt-free adaptation approach that eliminates the need for large annotated datasets by leveraging few-shot examples, which is used as memory guidance, enabling fully automated and anatomically-aware segmentation of complex structures.}
		
		\vspace{-0.5cm}
		\label{fig1}
	\end{figure*}
	
In this study, we propose a novel approach called Few-shot Adaptation of Training-frEe SAM (FATE-SAM), which leverages few-shot examples to adapt SAM2 for 3D medical image segmentation without model fine-tuning or manual prompts (Fig.~\ref{fig1}). FATE-SAM leverages SAM2's pre-trained memory mechanisms to integrate anatomical knowledge from few-shot examples while simultaneously ensuring volumetric consistency across a 3D volume, effectively eliminating the need for manual prompts. We evaluate the effectiveness of FATE-SAM on multiple 3D medical imaging datasets, including Computed Tomography (CT) and Magnetic Resonance Imaging (MRI) with different contrasts, covering various anatomical structures such as the knee, brain, heart, and abdomen. Experimental results show that FATE-SAM consistently achieves superior or comparable performance to competitive methods, including supervised learning models, zero-shot SAM approaches, and fine-tuned medical SAM variants.

The primary contributions of this study are as follows:

\begin{itemize}
    \item \textbf{Training-Free Adaptation:} We adapted SAM2 for medical image segmentation without requiring model fine-tuning.

    \item \textbf{Prompt-Free Segmentation:} We designed a few-shot adaptation pipeline that leverages a limited number of support examples to capture anatomical knowledge and eliminate the need for manual prompts.

    \item \textbf{Volumetric Consistency:} We introduced a volumetric mechanism to enforce 3D spatial consistency within a 3D medical volume.

    \item \textbf{Extensive Evaluation:} We evaluated FATE-SAM on 11 segmentation tasks across five diverse 3D medical imaging datasets, encompassing multiple imaging modalities and anatomical structures. The results demonstrate that FATE-SAM achieves robust performance matching or surpassing other competitive methods across various tasks.
    
\end{itemize}

	\section{Related Works}	
	\subsection{General Vision Foundation Models for Image Segmentation}
	
	General vision foundation models are pre-trained on massive, diverse datasets, enabling versatile visual representations adaptable to various tasks. Recent vision foundation models are built on Vision Transformer (ViT) blocks \cite{dosovitskiy2020image} and leverage self-supervised learning techniques \cite{gidaris2018unsupervised} for pre-training. Important vision foundation models include:  
	1) \textbf{DINO:} DINO \cite{caron2021emerging} and DINOv2 \cite{oquab2023dinov2} use self-distillation to learn features with unlabeled data.  
	2) \textbf{CLIP:} CLIP \cite{radford2021learning} aligns images with text descriptions through contrastive learning, enabling robust multi-modal understanding.  
	3) \textbf{Diffusion Models:} Diffusion models \cite{ho2020denoising} learn denoising processes to restore target images, excelling in general visual representation learning.  
	
	Adapting vision foundation models for segmentation has gained attention in recent research. Although those vision foundation models are not explicitly designed for segmentation tasks, they demonstrate emergent segmentation capabilities. For instance, DINO leverages its learned features to enable unsupervised segmentation by employing clustering techniques such as k-means or graph partitioning \cite{amir2021deep,simeoni2021localizing}, effectively identifying salient objects. CLIP can generate segmentation masks using its text encoder to classify mask proposals \cite{wang2025sclip}. Other approaches fine-tuned CLIP for segmentation with pixel-wise annotated data, such as DenseCLIP \cite{kwon2023probabilistic}. Additionally, parameter-efficient methods like ZegCLIP \cite{zhang2024language} aim to maintain the model's generalizability while improving segmentation performance. Diffusion models treat segmentation as a denoising process, progressively refining noisy inputs into precise segmentation masks \cite{chen2023generative,tian2024diffuse}.
	
    However, adapting general vision foundation models for dense segmentation tasks remains challenging. These models are inherently optimized for global context and high-level feature extraction, which limits their ability to achieve the fine-grained pixel-level precision essential for segmentation.

	\subsection{Dedicated Segmentation Foundation Models}
	
	To address the limitations of general vision foundation models in image segmentation, specialized models such as SAM have been developed \cite{kirillov2023segment}. SAM is trained on an extensive dataset of 1 billion masks from 11 million images, utilizing a promptable segmentation approach. This design enables SAM to achieve zero-shot generalization across a wide range of segmentation tasks \cite{kirillov2023segment}. By leveraging user-provided prompts, including points, bounding boxes, or masks that indicate the objects needed to be segmented, SAM can quickly adjust to various segmentation tasks without the need for additional fine-tuning. Unlike earlier universal models, which were constrained to fixed tasks such as semantic, instance, or panoptic segmentation with a closed vocabulary \cite{cheng2022masked,jain2023oneformer}, SAM’s promptable segmentation approach generates masks that directly correspond to the provided prompts instead of predicting specific classes. This flexibility makes SAM a powerful tool for tackling a diverse array of segmentation challenges.
	
	\subsection{SAM for Medical Image Segmentation}
	
	SAM has been explored for zero-shot segmentation across various medical imaging modalities, including CT \cite{roy2023sam}, MRI \cite{putz2024segment}, pathology \cite{deng2023segment}, and endoscopy \cite{wang2023sam}. While SAM performs comparably to state-of-the-art methods for some structures, it struggles with weak boundaries, low contrast, small sizes, and specific anatomy \cite{zhang2023towards}, highlighting the need for adaptations in complex scenarios.
	
    To enhance SAM's performance in medical image segmentation, various fine-tuning approaches have been developed. MedSAM \cite{ma2024segment} fine-tunes SAM using an extensive dataset of over one million medical image-mask pairs spanning 11 imaging modalities. To address the challenges associated with large-scale medical dataset requirements, parameter-efficient strategies have been introduced. For example, SAMed \cite{zhang2023customized} and SAMFE \cite{feng2023cheap} leverage LoRA modules to optimize either the image encoder or mask decoder. Similarly, SAM-Med2D \cite{cheng2023sam} incorporates adapter layers into SAM's architecture to improve performance with reduced data and computational requirements.
    
    Despite these advancements, the acquisition of pixel-wise annotated medical datasets remains a significant challenge. The annotation process is highly labor-intensive, requiring expert domain knowledge, and is further complicated by privacy concerns and restrictions on data sharing. Moreover, SAM's dependence on manual prompts, which require expert intervention, limits its practicality in real-world applications.
    
     In summary, the application of SAM in medical imaging faces two key challenges: the need for extensive fine-tuning and dependence on user-provided prompts. Addressing these limitations is crucial for enabling its seamless integration into clinical workflows.
    
	\section{Methods}
	
	\subsection{FATE-SAM Pipeline}
	
	FATE-SAM is developed utilizing pre-trained modules from SAM2, which was originally designed for time-series image (video) segmentation. In SAM2, several key network modules are combined to achieve high segmentation efficiency using an encoding-decoding approach, including: 
	1) \textbf{Image Encoder} $\mathcal{E}$ encodes an image into image embeddings.
	2) \textbf{Prompt Encoder} $\mathcal{PE}$ encodes the user-provided prompts into prompt embeddings.
	3) \textbf{Memory Encoder} $\mathcal{ME}$ encodes segmentation results from the past segmented frames into memory embeddings in the time-series process.
	4) \textbf{Memory Attention} $\mathcal{MA}$ uses the memory embeddings from the previous frames to guide the image embeddings in the current frame during the time-series process.
	5) \textbf{Mask Decoder} $\mathcal{D}$ generates segmentation masks by combining the embedded feature, prompt, and memory information.

    FATE-SAM leverages the pre-trained modules from SAM2 without model fine-tuning. Instead of relying on prompt embeddings driven by users' input, FATE-SAM incorporates few-shot adaptation, leveraging a limited number of support images and masks to provide anatomical priors, thereby eliminating the need for manual prompts.

\begin{figure*}[ht!]
		\centering
		\vspace{-0.7cm}
		\includegraphics[width=0.95\textwidth]{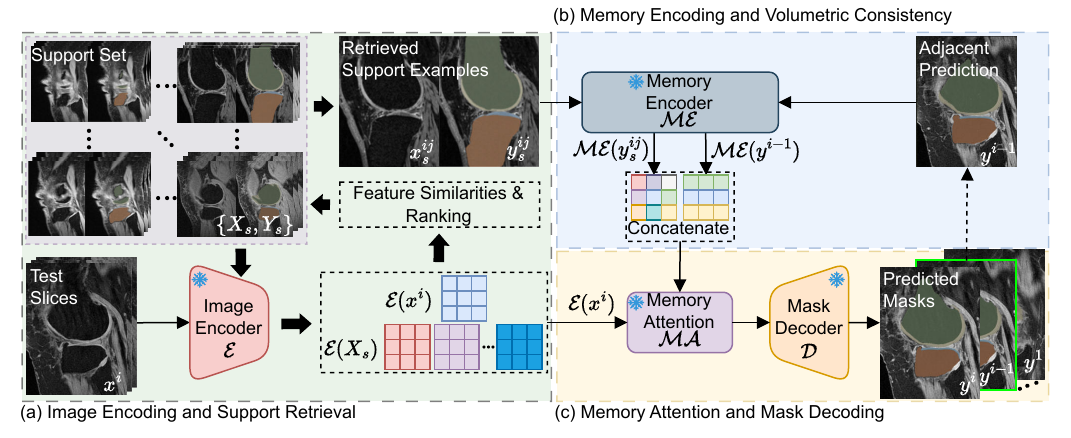}
		\vspace{-0.3cm}
		\caption{The Method Pipeline: 
			a) \textbf{Image Encoding and Support Retrieval}: The test slice and all support slices set are encoded into image embeddings, and the most similar support examples are retrieved by ranking the feature similarities between the support image embeddings and the test image embeddings. 
                b) \textbf{Memory Encoding and Volumetric Consistency}: Support examples are encoded into anatomical memory embeddings, while adjacent predictions are encoded into volumetric memory embeddings. These two types of memory are then fused to create unified memory embeddings, integrating both anatomical knowledge and volumetric consistency.
                c) \textbf{Memory Attention and Mask Decoding}: The unified memory embeddings are integrated into the test image embeddings through memory attention. This enriched representation guides the segmentation process, enabling the generation of accurate and coherent predictions.
            }
		\vspace{-0.6cm}
		\label{fig2}
	\end{figure*}

    The training-free and prompt-free FATE-SAM pipeline is implemented by reassembling the pre-trained modules $\mathcal{E}$, $\mathcal{ME}$, $\mathcal{MA}$, and $\mathcal{D}$ from SAM2, as illustrated in Fig.~\ref{fig2}. This framework introduces few-shot adaptation into the 3D medical image segmentation process through the following three major steps: 1) \textbf{Image Encoding and Support Retrieval}: The test slice and support slices are encoded using the Image Encoder $\mathcal{E}$ to generate image embeddings. Relevant support examples, including support slices and their corresponding masks, are retrieved from the support set based on the feature similarity between the test image embeddings and the support image embeddings. 2) \textbf{Memory Encoding and Volumetric Consistency}: The retrieved support examples are processed by the Memory Encoder $\mathcal{ME}$ to generate anatomical memory embeddings. Simultaneously, to ensure spatial coherence across slices, the segmentation prediction from the adjacent slice is also encoded by $\mathcal{ME}$ to produce volumetric memory embeddings. These two types of memory are fused to create unified memory embeddings that incorporate both anatomical knowledge and volumetric consistency. 3) \textbf{Memory Attention and Mask Decoding}: The unified memory embeddings are utilized to guide the test image embeddings via Memory Attention $\mathcal{MA}$. The final segmentation results are then generated by the Mask Decoder $\mathcal{D}$, which integrates the memory-guided test image embeddings into predictions.

    The following sections provide detailed descriptions of the three major steps in FATE-SAM. To formulate the problem, we consider the task of segmenting a 3D volume $x = \{x^1, x^2, \dots, x^n\}$, which consists of $n$ slices. Additionally, a support set $\{X_s, Y_s\}$ is provided, where $X_s = \{x_{s1}, x_{s2}, \dots, x_{sl}\}$ represents $l$ support volumes, and $Y_s = \{y_{s1}, y_{s2}, \dots, y_{sl}\}$ contains the corresponding support masks. The segmentation task is defined as a mapping function $y = \text{FATE}(x, \{X_s, Y_s\})$, which predicts the segmentation masks $y = \{y^1, y^2, \dots, y^n\}$ for all slices in the 3D volume $x$ using the guidance provided by the support set $\{X_s, Y_s\}$.

    \subsubsection{Image Encoding and Support Retrieval} 
In this step, relevant support examples are retrieved from the support set by evaluating similarities between the test slice and all support slices.

Before initializing the segmentation process, all slices in the support set $X_s$ are encoded into a library of support image embeddings, represented as $F_s = \mathcal{E}(X_s)$, using the pre-trained Image Encoder $\mathcal{E}$. Specifically, a Hiera ViT \cite{ryali2023hiera} pre-trained with Masked AutoEncoder (MAE) \cite{he2022masked} is used to extract these embeddings.

The 3D medical image segmentation process starts with an initial test slice $x^i$, where $1 \leq i \leq n$. The image embeddings for the test slice are encoded as $f^i = \mathcal{E}(x^i)$. To determine the similarity between $f^i$ and each embeddings in $F_s$, cosine similarity \cite{singhal2001modern} is calculated as:
\begin{equation}
    Sim(F_s, f^i) = \frac{f^i \cdot F_s}{||f^i|| \cdot ||F_s||}.
    \label{eq1:cosim}
\end{equation}

Based on the similarity scores, a ranked list of support image embeddings is generated. The top-$j$ most similar support image embeddings $f_s^{ij}$ and their corresponding masks $y_s^{ij}$ are selected as support examples:
\begin{equation}
    \{f_s^{ij}, y_s^{ij}\} = 
    \operatorname*{arg\,max}_{\substack{f_s^{ij} \in F_s \\ y_s^{ij} \in Y_s}} 
    Sim(F_s, f^i).
    \label{eq:retrival}
\end{equation}

These selected support examples $\{f_s^{ij}, y_s^{ij}\}$ may come from different support volumes, providing anatomical information into the segmentation process.

\subsubsection{Memory Encoding and Volumetric Consistency}

This step encodes anatomical information from selected support examples $\{f_s^{ij}, y_s^{ij}\}$ as anatomical memory while incorporating adjacent segmented slices as volumetric memory to ensure accurate and coherent segmentation across the 3D volume.

Specifically, for the anatomical memory, the selected support examples $\{f_s^{ij}, y_s^{ij}\}$ are processed by the Memory Encoder $\mathcal{ME}$, which consists of downsampling and pre-trained convolutional modules \cite{zhu2024medical}, to generate anatomical memory embeddings:
\begin{equation}
    \hat{f}_s^{ij} = \mathcal{ME}(y_s^{ij}) + f_s^{ij}.
    \label{eq:ME_fusion}
\end{equation}

For the volumetric memory, the image embeddings $f^{i-1}$ from the adjacent slice and its predicted mask $y^{i-1}$ are fed into the same Memory Encoder $\mathcal{ME}$ to produce volumetric memory embeddings:
\begin{equation}
    \tilde{f}^{i-1} = \mathcal{ME}(y^{i-1}) + f^{i-1}.
\end{equation}

The anatomical memory $\hat{f}_s^{ij}$ and volumetric memory $\tilde{f}^{i-1}$ are then fused via concatenation to construct unified memory embeddings:
\begin{equation}
    \dot{f}_s^i = [\hat{f}_s^{ij}, \tilde{f}^{i-1}].
    \label{eq8:featurefusion}
\end{equation}

Notably, $\tilde{f}^{i-1}$ is set to zero if $y^{i-1}$ is not available, such as when segmenting the initial slice.

\subsubsection{Memory Attention and Mask Decoding}

In this step, the anatomical knowledge and volumetric consistency contained in $\dot{f}^i_s$ are used to guide the test slice image embeddings $f^i$, enabling the generation of segmentation mask predictions.

Specifically, $\dot{f}^i_s$ and $f^i$ are processed by the Memory Attention module $\mathcal{MA}$, which combines the test image embeddings with the unified memory embeddings:
\begin{equation}
    \dot{f}^i = \mathcal{MA}(\dot{f}_s^i, f^i) = \mathcal{CA}(\dot{f}_s^i, \mathcal{SA}(f^i)).
    \label{eq:fused_memory}
\end{equation}
Here, $\mathcal{SA}$ and $\mathcal{CA}$ represent self-attention and cross-attention \cite{dao2023flashattention}, respectively, and are defined as follows:
\begin{equation}
    \mathcal{SA}(v) = \textit{softmax}\left(\frac{QK^T}{\sqrt{d}}\right)V + V,
    \label{eq4:SA}
\end{equation}
\begin{equation}
    \mathcal{CA}(v_1, v_2) = \textit{softmax}\left(\frac{Q_1K_2^T}{\sqrt{d}}\right)V_2 + V_2,
    \label{eq5:CA}
\end{equation}
where $Q$, $K$, and $V$ denote the query, key, and value projections of the input embeddings. For input embeddings $v$, these projections are computed as $Q = vW_q$, $K = vW_k$, and $V = vW_v$, with $W_q$, $W_k$, and $W_v$ representing pre-trained weight matrices from SAM2, and $d$ denoting the embedding dimensionality.

The resulting memory-guided embeddings $\dot{f}^i$ integrate anatomical knowledge and volumetric consistency, enriching the test slice representation. The final segmentation mask is generated by the Mask Decoder $\mathcal{D}$ as:
\begin{equation}
    y^i = \mathcal{D}(\dot{f}^i).
    \label{eq9:VolumeDecoder}
\end{equation}

For multi-object segmentation within a slice, the image embeddings generated by the Image Encoder $\mathcal{E}$ are shared across all objects. However, other modules, including the Memory Encoder $\mathcal{ME}$, Memory Attention $\mathcal{MA}$, and Mask Decoder $\mathcal{D}$ operate separately for each object. 

The above steps are propagated in both forward and backward directions until segmentation masks are generated for all slices in the 3D volume.

	\subsection{Experiments Design}
	To evaluate the effectiveness of FATE-SAM, we conducted experiments on diverse datasets benchmarking its performance against competitive methods. Furthermore, we performed ablation studies to analyze and validate the optimal configuration of FATE-SAM.
	
    \subsubsection{Datasets}
    
    We evaluated FATE-SAM on five publicly available 3D medical image datasets, which encompass 11 segmentation tasks across both CT and MRI scans with varying contrasts. These datasets include various anatomical structures, such as the knee, heart, brain, and abdomen, representing a total of 34 different anatomical objects. In total, we utilized 1,920 publicly available 3D medical volumes, which consist of 142,814 slices and their corresponding ground truth masks. A detailed summary of the datasets is provided in Table~\ref{table:datasets}. Within all the datasets, 10\% of the image volumes were designated as the support set, while the remaining 90\% were used for testing.
	
	\subsubsection{Comparison Studies}
    We compared FATE-SAM with \textbf{Supervised Methods}, \textbf{Zero-Shot SAM Methods}, and \textbf{Fine-Tuned Medical SAM Methods}. These methods rely on fine-tuning and/or manual prompts, as summarized in Table~\ref{table:comparison_settings}. In contrast, FATE-SAM eliminates the need for both fine-tuning and manual prompts.
    
	\textbf{Supervised Methods} included U-Net \cite{ronneberger2015u}, UNETR \cite{hatamizadeh2022unetr}, and SwinUNETR \cite{hatamizadeh2021swin}, which require task-specific training on pixel-wise annotated datasets. To enable inference, the same support set used in our method was used for their training set.
	
    \textbf{Zero-Shot SAM Methods} used frozen weights and random point prompts generated from ground truth masks. Since SAM is semantic-agnostic, individual positive points were created for each object. The final multi-object segmentation map was then generated by combining the segmentation results of all objects.
    
    \textbf{Fine-Tuned Medical SAM Methods} includes MedSAM \cite{ma2024segment}, SAM-Med3D \cite{wang2024sammed3dgeneralpurposesegmentationmodels}, and MedSAM2 \cite{zhu2024medical}. MedSAM and SAM-Med3D were fine-tuned on large-scale medical datasets enabling zero-shot inference for medical images, with their fine-tuned weights frozen during inference. MedSAM2, on the other hand, was adapted by fine-tuning on specific datasets, utilizing the same training data as the supervised methods. Additionally, these methods rely on manual prompts during inference. The specific prompt types integrated into their fine-tuning process, such as points or bounding boxes, are also generated from ground truth masks and provided during inference.

	\begin{table}[t!]
		\centering
		\vspace{-0.7cm}
		\caption{Medical image segmentation datasets used for evaluation. Each row indicates a segmentation task which may contain multiple segmentation objects.}
		\resizebox{0.48\textwidth}{!}{%
        \begin{tabular}{cccccc}
        \hline
        Dataset & Anatomy & \# of Objects & Modality & \begin{tabular}[c]{@{}c@{}}Support/\\ Test Volumes\end{tabular} & \begin{tabular}[c]{@{}c@{}}Support/\\ Test Slices\end{tabular} \\ \hline
        SKI10 \cite{heimann2010segmentation} & Knee & 4 & MRI & 10/90 & 1,088/9,786 \\ \hline
        ACDC \cite{bernard2018deep} & Heart & 3 & MRI & 20/180 & 371/3,408 \\ \hline
        BTCV \cite{landman2015miccai} & Abdominal & 13 & CT & 3/27 & 203/1,699 \\ \hline
        BraTS 2017 \cite{bakas2017advancing} & Brain & 3 & MRI & 49/435 & 3,396/30,359 \\ \hline
        \multirow{7}{*}{MSD \cite{antonelli2022medical}} & Hippocampus & 2 & MRI & 26/234 & 937/8,333 \\
         & Prostate & 2 & MRI & 3/29 & 70/532 \\
         & Lung & 1 & CT & 6/57 & 1,902/15,755 \\
         & Pancreas & 2 & CT & 28/253 & 3,094/23,625 \\
         & Hepatic Vessel & 2 & CT & 30/273 & 2,329/18,791 \\
         & Spleen & 1 & CT & 4/37 & 464/3,186 \\
         & Colon & 1 & CT & 13/113 & 1,301/12,185 \\ \hline
        \end{tabular}%
		}
		\vspace{-0.6cm}
		\label{table:datasets}
	\end{table}

	\subsubsection{Ablation Studies}
	We utilize the SKI10 dataset \cite{heimann2010segmentation} to examine how different configurations of FATE-SAM affect segmentation performance. The SKI10 dataset includes 3D knee MRI scans with delineated four anatomical structures: the femur bone, the tibia bone, the femoral cartilage, and the tibial cartilage. Generally, the femur and tibia bones are easier to segment due to their larger volume and rounded shapes. In contrast, segmenting the femoral and tibial cartilage is more challenging due to their smaller volume, thin structures, and curved shapes.

    	\begin{table}[t!]
		\centering
		\vspace{-0.7cm}
		\caption{Compared methods along with their respective settings and requirements for a single volume segmentation inference.}
		\resizebox{0.48\textwidth}{!}{%
			\begin{tabular}{ccccc}
				\hline
				Methods &
				\begin{tabular}[c]{@{}c@{}}General Image\\ Pretrained\end{tabular} &
				\begin{tabular}[c]{@{}c@{}}Medical Image\\ Fine-Tuning\end{tabular} &
				\begin{tabular}[c]{@{}c@{}}Manual\\ Prompt\end{tabular} &
				\begin{tabular}[c]{@{}c@{}}Dataset-Specific\\ Fine-Tuning\end{tabular} \\ \hline
				\multicolumn{5}{c}{Supervised Methods}                                                                                                              \\ \hline
				UNet \cite{ronneberger2015u}                                      & $\times$  & $\times$  & $\times$                                                            & \checkmark \\
				UNETR \cite{hatamizadeh2022unetr}                                 & $\times$  & $\times$  & $\times$                                                            & \checkmark \\
				Swin-UNETR \cite{hatamizadeh2021swin}                             & $\times$  & $\times$  & $\times$                                                            & \checkmark \\ \hline
				
				\multicolumn{5}{c}{Zero-Shot SAMs}                                                                                                               \\ \hline
				SAM \cite{kirillov2023segment}                                    & \checkmark & $\times$  & random points                                                    & $\times$  \\
				SAM2 \cite{ravi2024sam}                                           & \checkmark & $\times$  & random points                                                    & $\times$  \\
				\hline
				
				\multicolumn{5}{c}{Fine-Tuned Medical SAMs}                                                                                                                   \\ \hline
				MedSAM \cite{ma2024segment}                                       & \checkmark & \checkmark & bounding boxes & $\times$  \\
				
				MedSAM2 \cite{zhu2024medical}                                     & \checkmark & $\times$  & bounding boxes & \checkmark \\

				SAM-Med3D \cite{wang2024sammed3dgeneralpurposesegmentationmodels} & \checkmark & \checkmark & random points                                                    & $\times$  \\
				
				\hline
				Ours                                                              & $\checkmark$  & $\times$  & $\times$                                                            & $\times$ \\ \hline
			\end{tabular}%
		}
		\vspace{-0.6cm}
		\label{table:comparison_settings}
	\end{table}
    
	\textbf{i) Size of Support Set and Support Examples:} The size of the support set, represented by the number of support volumes $l$, affects the likelihood of retrieving well-matched examples. The number of support examples $j$ (as defined in Eq.~\ref{eq:retrival}) determines the richness of anatomical knowledge available to guide segmentation. Specifically, we investigated how varying $l$ (1\%–10\% of the total volumes) and $j$ (1–5 support examples) influences segmentation performance.
	
    \textbf{ii) Similarity Metrics Selection:} The choice of similarity metric in the support retrieval process (Eq.~\ref{eq:retrival}) plays an important role in guiding segmentation. To assess its impact, we evaluated FATE-SAM with various similarity metrics for support example retrieval. Specifically, we tested both image-level metrics, including Mean Square Error (MSE) \cite{wang2009mean} and Normalized Cross-Correlation (NCC) \cite{zhao2006image}, as well as feature-level metrics, such as Cosine Similarity (CS) \cite{singhal2001modern}, Manhattan Distance (MD) \cite{deza2009encyclopedia}, Euclidean Distance (ED) \cite{deza2009encyclopedia}, and Pearson Correlation Coefficient (PCC) \cite{cohen2009pearson}.

\textbf{iii) Prompts Comparison}: Our framework leverages few-shot adaptation to implicitly generate prompts from support examples for segmentation, eliminating the need for manual intervention. To assess the effectiveness of this implicit prompting, we compared it with manual prompts used in the original SAM2, including random points, bounding boxes, and masks. Random points and bounding boxes were generated from the ground truth masks, while mask prompts were derived from the same support masks utilized in our method.

    		\begin{figure*}[t!]
		\centering
		\vspace{-0.7cm}
		\resizebox{0.95\textwidth}{!}{%
			\includegraphics[width=\textwidth]{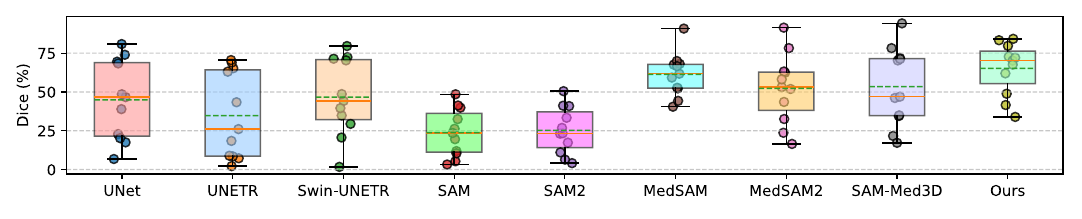}
		}
		\vspace{-0.5cm}
		\caption{Box plots comparison of Dice scores (\%) across 8 competitive methods and our method, evaluated on 11 tasks of interest. Each point indicates the average Dice score on a task. Each box represents the interquartile range (IQR) of the Dice scores for a specific method, with the horizontal line indicating the median. The whiskers extend to the minimum and maximum values within 1.5 times the IQR.}
		\label{figure:over_avg_compare}
	\end{figure*}

	\begin{table*}[htbp!]
		\renewcommand{\arraystretch}{1.1}
		\centering
		\vspace{-0.3cm}
            \caption{Segmentation performance comparison across multiple medical image datasets and tasks. Results are reported as the average Dice score $\pm$ standard deviation (unit: \%) over different objects in each task. The highest average Dice score in each column is highlighted in \textbf{bold}.}
            
		\centering
		
		\resizebox{\textwidth}{!}{%
			\begin{tabular}{c|ccccccccc}
				\hline
				\multirow{2}{*}{Methods/Datasets} & \multicolumn{5}{c|}{SKI10 \cite{heimann2010segmentation}} & \multicolumn{4}{c}{ACDC \cite{bernard2018deep}} \\ \cline{2-10} 
				& Femur Bone & Femoral Cartilage & Tibia Bone & Tibial Cartilage & \multicolumn{1}{c|}{Avg.} & Left Ventricle & Right Ventricle & Myocardium & Avg. \\ \hline
				UNet \cite{ronneberger2015u} & 87.13 ± 13.84 & 52.20 ± 21.77 & 86.99 ± 17.26 & 51.63 ± 20.04 & \multicolumn{1}{c|}{69.49} & 82.40 ± 21.43 & 69.17 ± 21.80 & 70.27 ± 20.58 & 73.95 \\
				UNETR \cite{hatamizadeh2022unetr} & 84.58 ± 13.64 & 46.00 ± 26.30 & 82.43 ± 16.88 & 48.04 ± 25.40 & \multicolumn{1}{c|}{65.26} & 73.88 ± 24.62 & 54.23 ± 25.84 & 61.04 ± 20.02 & 63.05 \\
				Swin-UNETR \cite{hatamizadeh2021swin} & 87.45 ± 13.33 & 54.46 ± 22.20 & 87.60 ± 17.03 & 56.06 ± 18.79 & \multicolumn{1}{c|}{71.39} & 80.58 ± 24.68 & 65.52 ± 24.76 & 71.16 ± 22.66 & 72.42 \\ \hline
				SAM \cite{kirillov2023segment} & 89.55 ± 11.60 & 7.32 ± 3.50 & 91.70 ± 14.60 & 5.37 ± 2.65 & \multicolumn{1}{c|}{48.49} & 69.27 ± 27.77 & 31.72 ± 30.18 & 18.61 ± 17.97 & 39.87 \\
				SAM2 \cite{zhu2024medical} & 68.58 ± 12.70 & 18.86 ± 7.90 & 64.12 ± 16.50 & 12.04 ± 1.85 & \multicolumn{1}{c|}{40.90} & 64.16 ± 27.10 & 41.44 ± 26.38 & 45.86 ± 18.85 & 50.49 \\ \hline
				MedSAM \cite{ma2024segment} & 89.14 ± 11.16 & 7.79 ± 13.46 & 90.37 ± 7.84 & 21.86 ± 19.44 & \multicolumn{1}{c|}{52.29} & \textbf{89.32 ± 11.09} & \textbf{87.10 ± 12.92} & 26.36 ± 27.33 & 67.59 \\
				MedSAM2 \cite{zhu2024medical} & 90.89 ± 10.40 & 26.69 ± 14.60 & 84.98 ± 16.50 & 47.19 ± 16.50 & \multicolumn{1}{c|}{62.44} & 46.58 ± 35.62 & 71.04 ± 22.80 & 71.96 ± 17.77 & 63.19 \\
				SAM-Med3D \cite{wang2024sammed3dgeneralpurposesegmentationmodels} & 57.72 ± 17.50 & 6.81 ± 2.75 & 65.90 ± 18.25 & 7.54 ± 4.69 & \multicolumn{1}{c|}{34.49} & 68.41 ± 15.14 & 43.63 ± 12.42 & 26.26 ± 15.40 & 46.10 \\ \hline
				Ours & \textbf{95.16 ± 9.74} & \textbf{66.81 ± 10.40} & \textbf{95.44 ± 7.80} & \textbf{61.48 ± 14.60} & \multicolumn{1}{c|}{\textbf{79.72}} & 88.41 ± 16.28 & 82.48 ± 21.94 & \textbf{79.00 ± 13.86} & \textbf{83.30} \\ \hline
				\multirow{2}{*}{Methods/Datasets} & \multicolumn{9}{c}{BTCV \cite{landman2015miccai}} \\ \cline{2-10} 
				& Spleen & Right kidney & Left Kidney & Gallbladder & Esophagus & Liver & Stomach & Aorta & Inferior Vena Cava \\ \hline
				UNet \cite{ronneberger2015u} & 54.85 ± 27.60 & 55.99 ± 30.36 & 48.70 ± 27.82 & 26.80 ± 26.55 & 26.72 ± 20.24 & 85.85 ± 8.70 & 31.66 ± 30.71 & 65.88 ± 20.24 & 48.66 ± 24.31 \\
				UNETR \cite{hatamizadeh2022unetr} & 8.65 ± 10.81 & 0.24 ± 0.65 & 2.14 ± 5.30 & 0.00 ± 0.00 & 0.00 ± 0.00 & 78.15 ± 10.34 & 10.44 ± 15.36 & 0.00 ± 0.00 & 0.00 ± 0.00 \\
				Swin-UNETR & 50.07 ± 29.31 & 47.78 ± 24.24 & 45.22 ± 27.18 & 35.58 ± 25.96 & 0.00 ± 0.00 & 80.88 ± 14.55 & 21.75 ± 17.32 & 64.54 ± 19.55 & 49.65 ± 15.28 \\ \hline
				SAM \cite{kirillov2023segment} & 15.53 ± 21.27 & 52.19 ± 40.17 & 62.19 ± 37.26 & 3.01 ± 8.56 & 4.12 ± 11.32 & 26.56 ± 18.70 & 19.01 ± 21.90 & 47.89 ± 38.93 & 16.14 ± 30.15 \\
				SAM2 \cite{zhu2024medical} & 20.24 ± 28.57 & 55.67 ± 43.35 & 66.94 ± 39.55 & 5.42 ± 17.24 & 11.55 ± 21.31 & 34.59 ± 27.11 & 28.35 ± 31.09 & 54.76 ± 41.80 & 16.42 ± 32.52 \\ \hline
				MedSAM \cite{ma2024segment} & 89.88 ± 26.25 & 91.13 ± 6.77 & 91.54 ± 4.65 & 73.31 ± 17.65 & \textbf{74.82 ± 7.95} & 94.78 ± 2.65 & 83.69 ± 9.65 & \textbf{88.10 ± 13.54} & 82.91 ± 12.55  \\ 
				MedSAM2 \cite{zhu2024medical} & 90.82 ± 10.88 & 85.03 ± 14.44 & 86.21 ± 13.90 & 66.32 ± 20.05 & 55.20 ± 20.58 & 89.27 ± 11.87 & 82.19 ± 12.47 & 86.79 ± 12.76 & 80.39 ± 16.00 \\
				SAM-Med3D \cite{wang2024sammed3dgeneralpurposesegmentationmodels} & \textbf{92.75 ± 3.27} & \textbf{91.28 ± 6.09} & \textbf{91.76 ± 5.02} & \textbf{75.13 ± 18.02} & 74.58 ± 5.36 & \textbf{94.88 ± 0.89} & \textbf{84.45 ± 8.68} & 82.96 ± 21.26 & \textbf{83.94 ± 7.98}\\ \hline
				Ours & 86.22 ± 19.42 & 90.99 ± 9.41 & 89.32 ± 11.86 & 69.09 ± 25.66 & 55.05 ± 23.94 & 82.17 ± 23.55 & 58.80 ± 33.87 & 87.38 ± 11.77 & 77.96 ± 20.91 \\ \hline
				\multirow{2}{*}{Methods/Datasets} & \multicolumn{5}{c|}{BTCV \cite{landman2015miccai}} & \multicolumn{4}{c}{BraTS 2017 \cite{bakas2017advancing}} \\ \cline{2-10} 
				& Portal Vein and Splenic Vein & Pancreas & Right adreanal gland & Left Adrenal gland & \multicolumn{1}{c|}{Avg.} & Edema & Non-Enhancing Tumor & Enhancing tumor & Avg. \\ \hline
				UNet \cite{ronneberger2015u} & 39.49 ± 19.98 & 21.44 ± 23.86 & 0.25 ± 0.96 & 0.00 ± 0.00 & \multicolumn{1}{c|}{38.94} & 29.88 ± 22.83 & 16.15 ± 17.72 & 14.72 ± 18.99 & 20.25 \\
				UNETR \cite{hatamizadeh2022unetr} & 9.31 ± 8.37 & 1.64 ± 3.89 & 0.00 ± 0.00 & 0.00 ± 0.00 & \multicolumn{1}{c|}{8.51} & 28.92 ± 22.03 & 13.46 ± 15.91 & 13.04 ± 18.04 & 18.47 \\
				Swin-UNETR & 36.72 ± 19.65 & 20.90 ± 14.29 & 0.00 ± 0.00 & 0.00 ± 0.00 & \multicolumn{1}{c|}{34.85} & 30.47 ± 21.97 & 15.05 ± 17.08 & 16.44 ± 20.16 & 20.65 \\ \hline
				SAM \cite{kirillov2023segment} & 3.46 ± 10.33 & 3.42 ± 6.55 & 0.46 ± 1.43 & 0.59 ± 4.10 & \multicolumn{1}{c|}{19.58} & 13.64 ± 1.28 & 8.92 ± 2.23 & 8.85 ± 1.76 & 10.47 \\
				SAM2 \cite{zhu2024medical} & 5.59 ± 16.65 & 3.31 ± 6.32 & 0.49 ± 1.19 & 0.40 ± 0.58 & \multicolumn{1}{c|}{23.37} & 13.49 ± 1.28 & 9.57 ± 2.55 & 10.13 ± 2.29 & 11.06 \\ \hline
				MedSAM \cite{ma2024segment} & 49.04 ± 18.25 & \textbf{76.16 ± 4.63} & 59.91 ± 9.35 & 61.38 ± 10.65 & \multicolumn{1}{c|}{78.20} & \textbf{67.53 ± 13.04} & 30.80 ± 17.35 & 42.54 ± 22.51 & \textbf{46.96}  \\
				MedSAM2 \cite{zhu2024medical} & 50.36 ± 28.86 & 61.75 ± 20.37 & 30.14 ± 19.66 & 44.51 ± 18.80 & \multicolumn{1}{c|}{69.92} &  16.54 ± 9.95 & 15.62 ± 12.18 & 17.40 ± 12.93 & 16.52 \\
				SAM-Med3D \cite{wang2024sammed3dgeneralpurposesegmentationmodels} & 46.86 ± 16.34 & 75.68 ± 6.07 & \textbf{60.33 ± 9.23} & \textbf{62.29 ± 10.40} & \multicolumn{1}{c|}{\textbf{78.22}} & 47.31 ± 2.77 & 39.40 ± 3.83 & \textbf{46.07 ± 3.72} & 44.26 \\ \hline
				Ours & \textbf{51.76 ± 20.78} & 45.66 ± 25.23 & 40.57 ± 19.92 & 43.80 ± 23.80 & \multicolumn{1}{c|}{67.60} & 39.02 ± 5.00 & \textbf{41.65 ± 7.34} & 44.08 ± 7.24 & 41.58 \\ \hline
				\multirow{2}{*}{Methods/Datasets} & \multicolumn{3}{c|}{MSD Hippocampus \cite{antonelli2022medical}} & \multicolumn{3}{c|}{MSD Prostate \cite{antonelli2022medical}} & \multicolumn{3}{c}{MSD Pancreas \cite{antonelli2022medical}} \\ \cline{2-10} 
				& Anterior & Posterior & \multicolumn{1}{c|}{Avg.} & Peripheral Zone & Transition Zone & \multicolumn{1}{c|}{Avg.} & Pancreas & Cancer & Avg. \\ \hline
				UNet \cite{ronneberger2015u} & 71.66 ± 8.63 & 65.05 ± 11.04 & \multicolumn{1}{c|}{68.35} & 35.93 ± 17.29 & 57.43 ± 23.64 & \multicolumn{1}{c|}{46.68} & 13.63 ± 10.34 & 0.00 ± 0.00 & 6.81 \\
				UNETR \cite{hatamizadeh2022unetr} & 71.24 ± 9.68 & 66.64 ± 11.99 & \multicolumn{1}{c|}{68.94} & 33.79 ± 17.13 & 52.92 ± 24.30 & \multicolumn{1}{c|}{43.36} & 4.45 ± 5.62 & 0.00 ± 0.00 & 2.23 \\
				Swin-UNETR & 73.62 ± 7.83 & 67.05 ± 10.93 & \multicolumn{1}{c|}{70.33} & 30.38 ± 15.57 & 58.04 ± 24.51 & \multicolumn{1}{c|}{44.21} & 62.08 ± 17.93 & 17.02 ± 21.94 & 39.55 \\ \hline
				SAM \cite{kirillov2023segment} & 33.71 ± 25.49 & 31.24 ± 23.31 & \multicolumn{1}{c|}{32.48} & 28.73 ± 23.29 & 53.84 ± 30.49 & \multicolumn{1}{c|}{41.29} & 3.83 ± 7.51 & 2.72 ± 8.78 & 3.28 \\
				SAM2 \cite{zhu2024medical} & 24.40 ± 21.37 & 21.88 ± 18.94 & \multicolumn{1}{c|}{23.14} & 29.29 ± 23.37 & 52.98 ± 30.65 & \multicolumn{1}{c|}{41.14} & 3.76 ± 8.32 & 4.51 ± 16.01 & 4.14 \\ \hline
				MedSAM \cite{ma2024segment} & 70.26 ± 15.40 & 65.87 ± 16.14 & \multicolumn{1}{c|}{68.07} & 46.32 ± 21.41 & 72.19 ± 15.37 & \multicolumn{1}{c|}{59.26} & 60.24 ± 21.47 & 63.24 ± 19.76 & 61.74 \\
				MedSAM2 \cite{zhu2024medical} & 62.01 ± 8.89 & 44.21 ± 15.88 & \multicolumn{1}{c|}{53.11} & 50.79 ± 18.20 & 53.03 ± 22.11 & \multicolumn{1}{c|}{51.91} & 36.58 ± 17.31 & 50.54 ± 18.24 & 43.56 \\
				SAM-Med3D \cite{wang2024sammed3dgeneralpurposesegmentationmodels} & 26.97 ± 4.49 & 16.38 ± 3.02 & \multicolumn{1}{c|}{21.67} & 60.93 ± 13.74 & \textbf{79.60 ± 8.24} & \multicolumn{1}{c|}{70.27} & \textbf{77.78 ± 8.49} & \textbf{65.89 ± 19.22} & \textbf{71.84} \\ \hline
				Ours & \textbf{73.93 ± 19.41} & \textbf{71.58 ± 20.53} & \multicolumn{1}{c|}{\textbf{72.75}} & \textbf{65.68 ± 22.73} & 78.43 ± 20.66 & \multicolumn{1}{c|}{\textbf{72.06}} & 49.36 ± 27.92 & 48.35 ± 26.50 & 48.85 \\ \hline
				\multirow{2}{*}{Methods/Datasets} & \multicolumn{1}{c|}{MSD Lung \cite{antonelli2022medical}} & \multicolumn{3}{c|}{MSD Hepatic Vessel \cite{antonelli2022medical}} & \multicolumn{1}{c|}{MSD Spleen \cite{antonelli2022medical}} & \multicolumn{1}{c|}{MSD Colon \cite{antonelli2022medical}} & \multicolumn{3}{c}{\multirow{2}{*}{Total Avg.}} \\ \cline{2-7}
				& \multicolumn{1}{c|}{Cancer} & Vessel & Tumor & \multicolumn{1}{c|}{Avg.} & \multicolumn{1}{c|}{Spleen} & \multicolumn{1}{c|}{Colon Cancer} & \multicolumn{3}{c}{} \\ \hline
				UNet \cite{ronneberger2015u} & \multicolumn{1}{c|}{22.86 ± 23.43} & 54.14 ± 10.07 & 42.87 ± 27.55 & \multicolumn{1}{c|}{48.50} & \multicolumn{1}{c|}{80.83 ± 11.42} & \multicolumn{1}{c|}{17.52 ± 16.72} & \multicolumn{3}{c}{44.93} \\
				UNETR \cite{hatamizadeh2022unetr} & \multicolumn{1}{c|}{7.27 ± 12.55} & 43.00 ± 11.36 & 9.14 ± 14.21 & \multicolumn{1}{c|}{26.07} & \multicolumn{1}{c|}{70.51 ± 12.80} & \multicolumn{1}{c|}{8.92 ± 12.11} & \multicolumn{3}{c}{34.78} \\
				Swin-UNETR & \multicolumn{1}{c|}{29.45 ± 19.85} & \textbf{55.47 ± 11.42} & 41.77 ± 27.57 & \multicolumn{1}{c|}{\textbf{48.62}} & \multicolumn{1}{c|}{79.56 ± 19.23} & \multicolumn{1}{c|}{1.74 ± 2.17} & \multicolumn{3}{c}{46.62} \\ \hline
				SAM \cite{kirillov2023segment} & \multicolumn{1}{c|}{26.28 ± 33.32} & 2.86 ± 9.96 & 7.81 ± 13.31 & \multicolumn{1}{c|}{5.33} & \multicolumn{1}{c|}{23.56 ± 28.61} & \multicolumn{1}{c|}{11.90 ± 21.76} & \multicolumn{3}{c}{23.87} \\
				SAM2 \cite{zhu2024medical} & \multicolumn{1}{c|}{33.37 ± 37.47} & 3.07 ± 10.53 & 9.76 ± 17.71 & \multicolumn{1}{c|}{6.41} & \multicolumn{1}{c|}{26.85 ± 34.81} & \multicolumn{1}{c|}{17.35 ± 26.59} & \multicolumn{3}{c}{25.29} \\ \hline
				MedSAM \cite{ma2024segment} & \multicolumn{1}{c|}{52.70 ± 24.50} & 12.96 ± 5.65 & \textbf{68.04 ± 13.45} & \multicolumn{1}{c|}{40.50} & \multicolumn{1}{c|}{90.81 ± 10.37} & \multicolumn{1}{c|}{66.57 ± 20.51} & \multicolumn{3}{c}{61.25} \\
				MedSAM2 \cite{zhu2024medical} & \multicolumn{1}{c|}{23.71 ± 24.25} & 29.87 ± 13.46 & 35.27 ± 25.42 & \multicolumn{1}{c|}{32.57} & \multicolumn{1}{c|}{91.41 ± 3.56} & \multicolumn{1}{c|}{58.13 ± 18.83} & \multicolumn{3}{c}{52.25} \\
				SAM-Med3D \cite{wang2024sammed3dgeneralpurposesegmentationmodels} & \multicolumn{1}{c|}{17.23 ± 14.30} & 4.46 ± 1.84 & 65.39 ± 12.65 & \multicolumn{1}{c|}{34.93} & \multicolumn{1}{c|}{\textbf{94.17 ± 1.77}} & \multicolumn{1}{c|}{\textbf{71.21 ± 12.57}} & \multicolumn{3}{c}{53.37} \\ \hline
				Ours & \multicolumn{1}{c|}{\textbf{70.19 ± 18.02}} & 34.72 ± 20.03 & 33.11 ± 28.87 & \multicolumn{1}{c|}{33.91} & \multicolumn{1}{c|}{84.21 ± 21.41} & \multicolumn{1}{c|}{61.96 ± 28.55} & \multicolumn{3}{c}{\textbf{65.10}} \\ \hline
			\end{tabular}%
		}
		
		\vspace{-0.5cm}
		\label{table:comparison_result}
	\end{table*}
    
	\textbf{iv) Volumetric Consistency:} We evaluated the necessity of Volumetric Consistency by comparing segmentation performance with and without it (i.e., set $\tilde{f}^{i-1}\equiv0$ in Eq. \ref{eq8:featurefusion}).
	
	\textbf{v) Initial Slice Selection:} The 3D segmentation process of FATE-SAM can be initialized from any slice in a volume, and the choice of the initial slice may influence the propagation of segmentation accuracy across adjacent slices. To evaluate this, We analyzed the impact of starting from different initial slices (i.e., different values for $i$) on segmentation results. $i$ was varied across five positions within the volume: the first slice, the last slice, the 25\% (Q1) position, the 75\% (Q3) position, and the central slice.
	
	\textbf{vi) Pre-Trained Weights Selection}:	There are various pre-trained weights of SAM2 with different model sizes. To identify the most effective weights for our method, we conducted a comparative analysis of various SAM2 versions, assessing their impact on segmentation performance. The evaluated weights include \textit{SAM2\_tiny}, \textit{SAM2\_small}, \textit{SAM2\_base+}, and \textit{SAM2\_large}.

	\subsubsection{Evaluation Metric and Implementation Details}
	Segmentation performance was evaluated using the Dice score, which measures the similarity between predicted and ground truth masks, with a range of 0 to 1. All experiments were conducted on Rocky Linux 8.8 with an Intel Xeon Gold 6338 CPU @ 2.00GHz and an NVIDIA A100 GPU. The \textit{SAM2\_large} pre-trained weights were used and kept frozen during inference for FATE-SAM. No post-processing was applied to the segmentation results.

    \vspace{-0.3cm}
    \section{Results}	
    \subsection{Comparison Study Results}

    	\begin{figure*}[t!]
		\centering
        \vspace{-0.7cm}
		\resizebox{1.\textwidth}{!}{%
			\includegraphics[width=\textwidth]{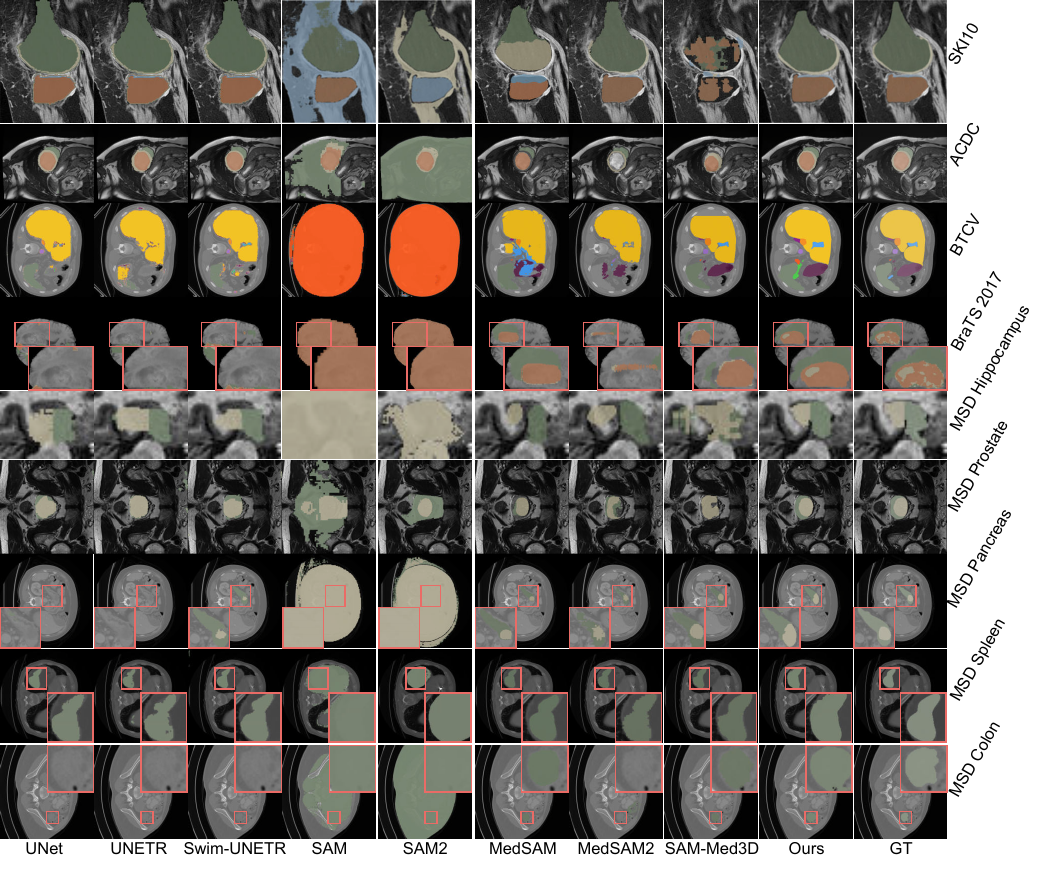}
		}
		\vspace{-0.9cm}
        \caption{Visual examples of segmentation results across datasets for different methods. Pink boxes enlarge the segmented objects for enhanced visualization. GT represents the ground truth masks.}
		\vspace{-0.6cm}
		\label{fig:2D_vis_result}
	\end{figure*}
    
Fig~\ref{figure:over_avg_compare} demonstrates that our method achieves the highest median Dice score along with a small interquartile range. This indicates both good accuracy and stability across diverse datasets. In comparison, supervised methods exhibit moderate median Dice scores but show broader variability, reflecting their instability when trained on limited data. Zero-shot SAM approaches yield lower Dice scores, highlighting their limited adaptability to medical segmentation tasks without fine-tuning. Meanwhile, fine-tuned medical SAM models perform well on specific datasets, but they lack robustness across a variety of tasks. Table~\ref{table:comparison_result} further provides detailed results of the segmentation performance of all methods across different datasets, tasks, and anatomical objects.
	
\textbf{SKI10 Dataset:} Our method achieved an average Dice score of 79.72\%, exceeding that of supervised methods, such as Swin-UNETR, which scored 71.39\%, and fine-tuned SAMs, like MedSAM2, which scored 62.44\%. As illustrated in Fig. ~\ref{fig:2D_vis_result} (row 1), most methods successfully segmented the femur (shown in green) and tibia (shown in brown). However, the supervised methods struggled with smooth boundaries for femoral cartilage (yellow) and tibial cartilage (blue), zero-shot SAMs failed, and the fine-tuned SAMs exhibited considerable variability. In contrast, our method produced smoother and anatomically accurate cartilage masks, achieving scores of 66.81\% for femoral cartilage and 61.48\% for tibial cartilage.
	
    \textbf{ACDC Dataset:} Our method achieved an average Dice score of 83.30\%, outperforming supervised methods such as UNet (73.95\%) and fine-tuned SAMs like MedSAM2 (67.59\%). As shown in Fig.\ref{fig:2D_vis_result} (row 2), our method demonstrated superior segmentation performance, particularly for the left ventricle (brown, 88.41\%) and the right ventricle (green, 87.10\%). The myocardium (yellow), however, presents significant challenges due to its thin, specific anatomical structure. In this case, supervised methods displayed moderate performance; for example, Swin-UNETR achieved a Dice score of 71.16\%. Zero-shot SAMs struggled to differentiate the myocardium from the right ventricle, highlighting their limited adaptability to medical domain knowledge. While fine-tuned SAMs showed improved performance, they exhibited high variability, with Dice scores ranging from 26.26\% to 71.96\%. In contrast, our method achieved a Dice score of 79.00\%, offering smoother and anatomically consistent segmentations.
	
	\textbf{BTCV Dataset:} Our method achieved an average Dice score of 67.60\%, behind fine-tuned SAMs (e.g., SAM-Med3D: 78.22\%). As shown in Fig.~\ref{fig:2D_vis_result} (row 3), SAM-Med3D excelled in segmenting abdominal structures such as the liver and spleen, benefiting from extensive fine-tuning on related medical datasets. These results highlight the advantages of fine-tuned models in leveraging domain-specific data to achieve high accuracy on anatomical structures. In contrast, supervised methods struggled (e.g., UNet: 38.94\%), largely due to limited training data and the inherent class imbalance in the BTCV dataset, which restricted their generalizability. Our method, without requiring fine-tuning, delivered competitive results, achieving Dice scores of 90.99\% for the right kidney and 87.38\% for the aorta, both comparable to fine-tuned models. For more challenging structures, such as the portal vein and splenic vein, characterized by small size and specific structure, our method outperformed fine-tuned SAMs. This demonstrates its ability to effectively capture complex anatomical features by leveraging few-shot examples.

\textbf{BraTS 2017 Dataset:} Our method achieved an average Dice score of 41.58\%, which outperformed supervised methods, such as UNet, which scored 20.65\%. Additionally, our performance was competitive when compared to fine-tuned SAM methods, like MedSAM, which achieved a Dice score of 46.96\%. As shown in Fig. \ref{fig:2D_vis_result} (row 4), our method demonstrated good segmentation performance across tumor subregions. For the Non-Enhancing Tumor (shown in orange), our method obtained a Dice score of 41.65\%. This surpassed the performance of SAM-Med3D, which scored 39.40\%, and outperformed supervised methods like UNet, which had a score of 16.15\%. Similarly, for the Enhancing Tumor (indicated in yellow), known for its heterogeneous textures and blurred edges, our method achieved a Dice score of 44.08\%. This score is close to that of SAM-Med3D at 46.07\%, while exceeding the performance of supervised methods, such as UNet, which scored 16.44\%.
	
\textbf{MSD Dataset:} Our method demonstrated strong generalization across diverse structures, as illustrated in Fig. \ref{fig:2D_vis_result} (row 5-9): 1) \textbf{Hippocampus:} Achieved a score of 72.75\%, surpassing UNETR (68.94\%) and Swin-UNETR (70.33\%), with smooth and accurate boundaries. 2) \textbf{Prostate:} Scored 72.06\%, featuring a strong performance in both the peripheral (65.68\%) and transition zones (78.43\%). 3) \textbf{Pancreas:} Produced anatomically consistent masks with both pancreas and pancreatic cancer. 4) \textbf{Spleen:} Delivered accurate and smooth delineations that were comparable to other methods. 5) \textbf{Colon:} Effectively managed variability in cancer segmentation, maintaining smooth and precise mask representations.
	
The comparison results demonstrate our method's adaptability and robustness, consistently outperforming or matching competitors across various datasets and anatomical structures.
	
	\subsection{Ablation Study Results}
	
\begin{figure}[t!]
    \centering
    \resizebox{0.48\textwidth}{!}{%
        \includegraphics[width=\textwidth]{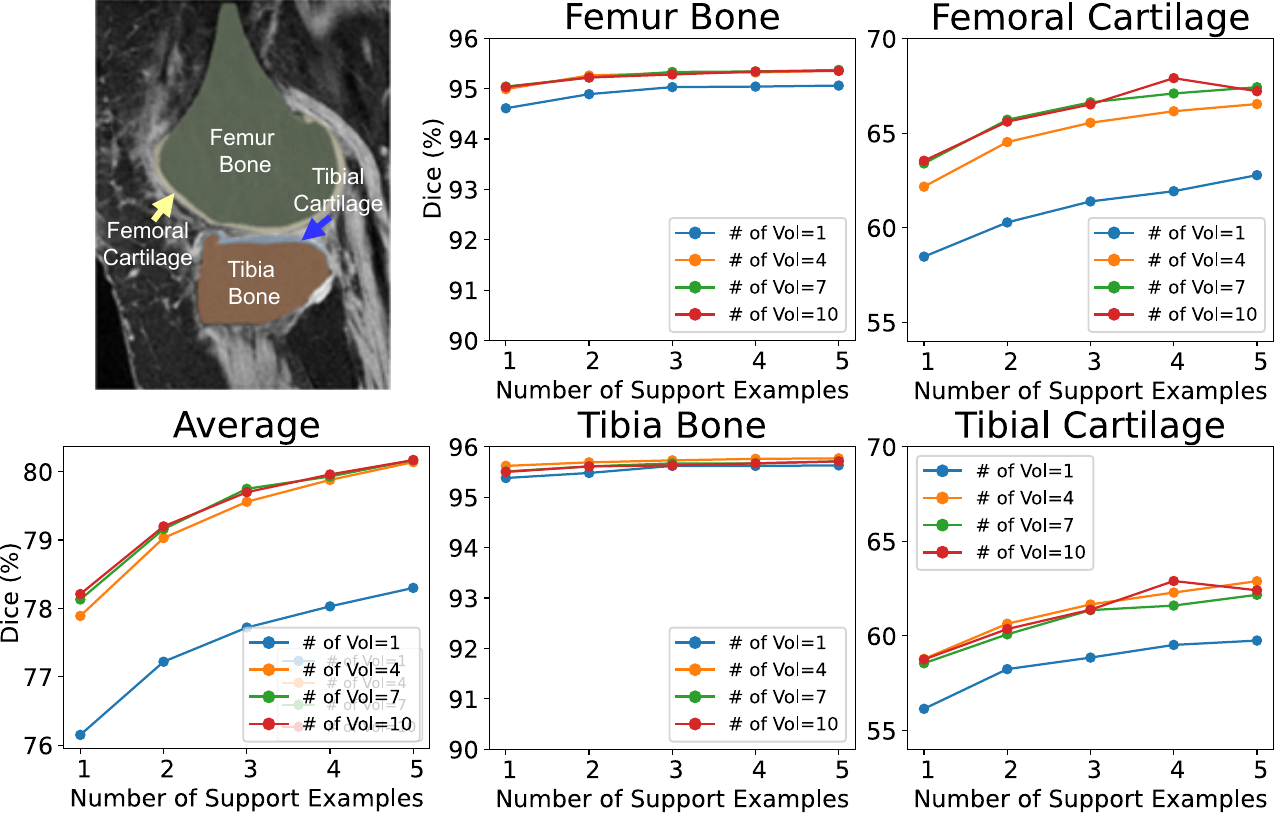}
    }
    \caption{Size of support set and support example ablation results on the SKI10 dataset. The top-left example illustrates the anatomical structures of the femur bone, tibia bone, femoral cartilage, and tibial cartilage. The plots display the segmentation performance of our method with varying sizes of the support set and the number of support examples. The horizontal axis indicates the number of support examples, while lines in different colors represent results for different sizes of the support set. Each sub-figure highlights the ablation performance for a specific anatomical structure.}
    \label{figure:few-shot-ablation}
\end{figure}

	In this section, we address key questions about the configuration of FATE-SAM through experimental results:
	
	\textbf{i) Size of Support Set and Support Examples:} Fig.~\ref{figure:few-shot-ablation} shows that increasing the support set size (1 to 10 volumes) and the number of support examples improves segmentation performance. For instance, the average Dice score increases from 78.03\% (1 volume) to 80.17\% (10 volumes) when using 5 support examples. This improvement is notable for challenging structures like femoral cartilage (62.78\% to 67.22\%) and tibial cartilage (59.75\% to 62.41\%), where greater support set diversity enhances generalization. Similarly, increasing support examples boosts performance for soft tissues, with femoral cartilage improving from 58.75\% (1 example) to 62.41\% (5 examples) using 10 volumes. For relatively easier femur bone, performance plateaus at 95\% accuracy with just 2 support examples.
	
	The best performance (80.17\%) is achieved with 10 volumes (10\% of the SKI10 dataset) and 5 support examples per slice. However, using more examples per slice significantly increases computational cost. Balancing efficiency and accuracy, we selected \textbf{10\% of total volumes} and \textbf{3 support examples} for FATE-SAM.
	
    \textbf{ii) Similarity Metrics Selection:} Table~\ref{table:similarity-metrics-ablation} demonstrates that feature-level retrieval consistently outperforms image-level retrieval. Image-level retrieval, relying on raw image intensities, has limited representational capacity for capturing complex structures. In contrast, feature-level retrieval utilizes the rich, high-level embeddings from the SAM2 encoder, enabling it to effectively capture structural details and semantic relationships.
    
    Among feature-level metrics, segmentation performance is comparable, with CS (Cosine Similarity) achieving the highest average Dice score of 79.72\%. Thus, we selected CS as the similarity metric for our method due to its simplicity and effectiveness.

    \begin{table}[t!]
		\centering
		\caption{Effect of Similarity Metrics. Results denote Dice (unit:\%) on the SKI10 dataset. 
		}
		\begin{tabular}{cccccc}
			\hline
			\multicolumn{1}{c|}{\begin{tabular}[c]{@{}c@{}}Similarity\\ Metrics\end{tabular}} & \begin{tabular}[c]{@{}c@{}}Femur\\ Bone\end{tabular} & \begin{tabular}[c]{@{}c@{}}Femoral\\ Cartilage\end{tabular} & \begin{tabular}[c]{@{}c@{}}Tibia\\ Bone\end{tabular} & \begin{tabular}[c]{@{}c@{}}Tibial\\ Cartilage\end{tabular} & Avg. \\ \hline
			\multicolumn{6}{c}{Image Level} \\ \hline
			\multicolumn{1}{c|}{MSE} & 94.75 & 63.75 & 95.00 & 58.11 & 77.90 \\
			\multicolumn{1}{c|}{NCC} & 95.13 & 64.51 & 95.17 & 59.71 & 78.63 \\ \hline
			\multicolumn{6}{c}{Feature Level} \\ \hline
			\multicolumn{1}{c|}{CS} & 95.16 & 66.81 & 95.44 & \textbf{61.48} & \textbf{79.72} \\
			\multicolumn{1}{c|}{MD} & 95.27 & 66.51 & 95.63 & 61.38 & 79.70 \\
			\multicolumn{1}{c|}{ED} & \textbf{95.29} & 66.33 & \textbf{95.65} & 61.31 & 79.64 \\
			\multicolumn{1}{c|}{PCC} & 95.18 & \textbf{66.89} & 95.40 & 59.48 & 79.24 \\ \hline
		\end{tabular}%
		\vspace{-0.3cm}
		\label{table:similarity-metrics-ablation}
	\end{table}

    	\begin{table}[t!]
		\centering
		\caption{Effect of prompts. Results denote Dice (unit:\%) on the SKI10 dataset.}
		\resizebox{0.48\textwidth}{!}{%
			\begin{tabular}{c|ccccc}
				\hline
				Prompt &
				\begin{tabular}[c]{@{}c@{}}Femur \\ Bone\end{tabular} &
				\begin{tabular}[c]{@{}c@{}}Femoral\\ Cartilage\end{tabular} &
				\begin{tabular}[c]{@{}c@{}}Tibia\\ Bone\end{tabular} &
				\begin{tabular}[c]{@{}c@{}}Tibial\\ Cartilage\end{tabular} &
				Avg. \\ \hline
				Random Points  & 68.58 & 18.86 & 64.12 & 12.04 & 40.90 \\
				Bounding Boxes   & \textbf{95.37} & 38.15 & \textbf{95.93} & 38.02 & 57.36 \\
				Masks          & 88.52 & 51.90 & 84.29 & 41.29 & 66.50 \\
				Few-shot (Ours)      & 95.16 & \textbf{66.81} & 95.44 & \textbf{61.48} & \textbf{79.72} \\ \hline
			\end{tabular}%
		}
		\vspace{-0.1cm}
		\label{table:prompt-ablation}
	\end{table}

    	\textbf{iii) Prompts Comparison}: Table~\ref{table:prompt-ablation} highlights the impact of different prompts on SAM2’s performance. Random point prompts result in lower segmentation (average Dice: 40.90\%), particularly for cartilage (e.g., 18.86\% for femoral cartilage, 12.04\% for tibial cartilage). Bounding box prompts improve performance (average Dice: 57.36\%), excelling in the tibia bone, but remain inadequate for cartilages (e.g., 38.15\% for femoral cartilage, 38.02\% for tibial cartilage). Mask prompts further enhance SAM2’s results (average Dice: 66.50\%), particularly for cartilage segmentation.

         Our approach achieves an average Dice score of 79.72\%, providing balanced improvements across both bone and cartilage segmentation, with scores of 66.81\% for femoral cartilage and 61.48\% for tibial cartilage. This reflects our few-shot adaptation can leverage anatomical knowledge from support examples effectively.
	
	\textbf{iv) Volumetric Consistency:} Table~\ref{table:3D_consistency} shows that incorporating volumetric consistency significantly improves segmentation performance, increasing the average Dice score from 70.06\% (without) to 79.72\% (with). The improvement is especially notable for cartilage segmentation, with femoral cartilage Dice score rising from 46.37\% to 66.81\% and tibial cartilage from 46.12\% to 61.48\%.
	
	These results highlight the importance of preserving structural continuity and leveraging spatial correlations.
	
	\textbf{v) Initial Slice Selection:} Table~\ref{table:initial_slice} shows that the initial slice choice has minimal impact on segmentation performance, with average Dice scores ranging from 79.21\% to 80.03\%. Slices around Q3 perform slightly better (80.03\%), demonstrating the method's robustness to initial slice selection.
	
	We chose to initialize FATE-SAM from the central slice, as it offers abundant anatomical structures and is intuitive for human interpretation.

        	\begin{table}[t!]
		\centering
		\caption{Effect of Volumetric Consistency. Results denote Dice (unit:\%) on the SKI10 dataset.}
		\resizebox{0.48\textwidth}{!}{%
			\begin{tabular}{c|ccccc}
				\hline
				\begin{tabular}[c]{@{}c@{}}Initial \\ Slice\end{tabular} & \begin{tabular}[c]{@{}c@{}}Femur \\ Bone\end{tabular} & \begin{tabular}[c]{@{}c@{}}Femoral\\ Cartilage\end{tabular} & \begin{tabular}[c]{@{}c@{}}Tibia\\ Bone\end{tabular} & \begin{tabular}[c]{@{}c@{}}Tibial\\ Cartilage\end{tabular} & Avg. \\ \hline
				w/o 3D consistency & 93.02 & 46.37 & 94.74 & 46.12 & 70.06 \\
				w/ 3D consistency & \textbf{95.16} & \textbf{66.81} & \textbf{95.44} & \textbf{61.48} & \textbf{79.72} \\ \hline
			\end{tabular}%
			}
		\vspace{-0.3cm}
		\label{table:3D_consistency}
	\end{table}
	
	\textbf{vi) Pre-Trained Weights Selection}: Table~\ref{table:sam2_weights} shows that our method achieves consistently high performance across SAM2 pre-trained weight variants, with average Dice scores ranging from 79.01\% to 79.72\%. \textit{SAM2\_large} achieves the highest score (79.72\%). With smaller weights, femur and tibia bones consistently perform well, while tibial cartilage scores slightly lower due to its complex anatomy and indistinct boundaries, but the differences are minimal, demonstrating robustness across model sizes. These findings indicate that FATE-SAM is effective regardless of the SAM2 variant, allowing model size to be chosen based on computational resource constraints rather than segmentation performance.
	
    In summary, the ablation study results highlight the robust performance of FATE-SAM across varying support sizes and confirm the effectiveness of the few-shot adaptation approach and volumetric consistency in enhancing segmentation performance. Additionally, FATE-SAM demonstrates insensitivity to the choice of similarity metrics for support retrieval, initial slice selection, and pre-trained weights.
    
	\begin{table}[t!]
		\centering
		\caption{Effect of initial slice. Q1/Q3 denote the slice at the 25\%/75\% of all slices. Results denote Dice (unit:\%) on the SKI10 dataset.}
			\begin{tabular}{c|ccccc}
				\hline
				\begin{tabular}[c]{@{}c@{}}Initial \\ Slice\end{tabular} &
				\begin{tabular}[c]{@{}c@{}}Femur \\ Bone\end{tabular} &
				\begin{tabular}[c]{@{}c@{}}Femoral\\ Cartilage\end{tabular} &
				\begin{tabular}[c]{@{}c@{}}Tibia\\ Bone\end{tabular} &
				\begin{tabular}[c]{@{}c@{}}Tibial\\ Cartilage\end{tabular} &
				Avg. \\ \hline
				First  & 95.15 & 65.04 & \textbf{95.75} & 60.92 & 79.21 \\
				Q1     & 95.18 & 66.29 & 95.61 & 61.67 & 79.69 \\
				Center & 95.16 & \textbf{66.81} & 95.44 & 61.48 & 79.72 \\
				Q3     & \textbf{95.31} & 66.76 & 95.62 & 62.42 & \textbf{80.03} \\
				Last   & 95.20 & 66.21 & 95.69 & \textbf{62.51} & 79.90 \\ \hline
			\end{tabular}%
			\vspace{-0.2cm}
		\label{table:initial_slice}
	\end{table}

	\begin{table}[t!]
		\centering
		\vspace{-0.2cm}
		\caption{Effect of SAM2 pretrained weights. RESULTS DENOTE Dice (UNIT:\%) on The SKI10 Dataset}
		\resizebox{0.48\textwidth}{!}{%
			\begin{tabular}{c|cccccc}
				\hline
				\begin{tabular}[c]{@{}c@{}}SAM2\\ Weights\end{tabular} &
				\begin{tabular}[c]{@{}c@{}}Size\\ (MB)\end{tabular} &
				\begin{tabular}[c]{@{}c@{}}Femur\\ Bone\end{tabular} &
				\begin{tabular}[c]{@{}c@{}}Femoral\\ Cartilage\end{tabular} &
				\begin{tabular}[c]{@{}c@{}}Tibia\\ Bone\end{tabular} &
				\begin{tabular}[c]{@{}c@{}}Tibial\\ Cartilage\end{tabular} &
				Avg. \\ \hline
				SAM2\_tiny &38.9 & 95.37 & 66.22 & \textbf{95.87} & 59.00 & 79.11 \\
				SAM2\_small &46 & 95.36& 66.14& 95.60& 60.81& 79.48\\
				SAM2\_base+ &80.8 & \textbf{95.39} & 64.50 & 95.85 & 60.32 & 79.01 \\
				SAM2\_large &224.4 & 95.16 & \textbf{66.81}& 95.44 & \textbf{61.48} & \textbf{79.72} \\ \hline
			\end{tabular}%
		}
		\vspace{-0.5cm}
		\label{table:sam2_weights}
	\end{table}
	
	\section{Discussion and Conclusion}
In this study, we introduced Few-shot Adaptation of Training-frEe SAM (FATE-SAM), a novel approach for 3D medical image segmentation that does not require fine-tuning or manual prompts. By utilizing few-shot examples and ensuring volumetric consistency, FATE-SAM achieves robust and accurate segmentation across a variety of imaging modalities and anatomical structures. Experiments conducted on multiple datasets demonstrated its competitive performance, highlighting its potential for clinical applications. Additionally, ablation studies affirmed its robustness and effectiveness.

Despite its strengths, FATE-SAM has some limitations. The inference process can be computationally intensive due to the retrieval of support examples. It also struggles with small or ambiguous structures, such as pancreatic cancer. Furthermore, FATE-SAM lacks post-processing capabilities, which may result in minor inaccuracies in complex cases. Future work could address these challenges by:
1. Optimizing computational efficiency for faster inference;
2. Enhancing segmentation of complex structures through advanced attention mechanisms with minimal module fine-tuning on a limited amount of data;
3. Incorporating post-processing techniques with anatomical priors to improve accuracy.

In summary, FATE-SAM represents a significant advancement in adapting foundation models for medical imaging, offering an efficient and accessible solution for clinical segmentation tasks.
	
	\bibliographystyle{IEEEtran}  
	\bibliography{references}     
	
\end{document}